\documentclass[11pt]{article}
\usepackage[margin=1in]{geometry}
\usepackage{amsmath,amssymb}
\usepackage{graphicx}
\usepackage{booktabs}
\usepackage{multirow}
\usepackage{threeparttable}
\usepackage{algorithm}
\usepackage{algpseudocode}
\usepackage[numbers,sort&compress]{natbib}
\usepackage[colorlinks, urlcolor=blue]{hyperref}

\newcommand{\best}[1]{\textbf{#1}}
\newcommand{\second}[1]{\underline{#1}}

\begin{document}

\title{RIGS-Refiner: Risk-Guided Recursive Refinement in Prediction Space for Colonoscopy Polyp Segmentation}

\author{
Jiachi Zhang$^{1,2}$, Zhuoyu Wu$^{3}$, and Wenqi Fang$^{1}$\\
\small $^1$Shenzhen Institutes of Advanced Technology, Chinese Academy of Sciences, Shenzhen, China\\
\small $^2$South China University of Technology, Guangzhou, China\\
\small $^3$CyPhi($\Psi\Phi$) AI Research Lab, School of IT, Monash University Malaysia\\
\small Corresponding authors: Zhuoyu Wu and Wenqi Fang\\
\small \texttt{wq.fang@siat.ac.cn}, \texttt{zhuoyu.wu@monash.edu}
}
\date{}

\maketitle

\begin{abstract}
Post-refinement can improve colonoscopy segmentation after host inference, but many designs still rely on extra correction heads or multi-stage pipelines with non-negligible parameter or computational cost.
For polyp segmentation, host predictions are often already reasonable globally, with remaining errors clustered around ambiguous boundaries and difficult local structures.
These residual errors matter in colonoscopy images because useful masks need correct lesion coverage and clean contour delineation across subtle mucosal transitions.
This setting favors selective local repair in prediction space over reprocessing the entire mask.
We therefore propose \emph{RIGS-Refiner}, a lightweight post-refinement plugin for risk-guided recursive refinement in prediction space.
Starting from a frozen host anchor prediction, RIGS-Refiner extracts lightweight image priors and prediction cues, applies risk-guided update, and writes back residual corrections through a shared recursive cell.
The module adds only $+519$ parameters and $+0.631$ GFLOPs, keeping the refinement path compact for deployment.
Experiments use Kvasir-SEG for training and Kvasir, ClinicDB, ColonDB, and ETIS for evaluation under two frozen hosts, namely PraNet and SegFormer-B0.
Results show consistent gains on both hosts and a favorable efficiency--accuracy trade-off against representative post-refinement methods.
Code is available at \url{https://github.com/tyui99/RIGS-Refiner}.
\end{abstract}

\noindent\textbf{Keywords:} Polyp segmentation, post-refinement, recursive correction, prediction space, lightweight segmentation, biomedical image processing.

\section{Introduction}
Accurate polyp segmentation is important for computer-aided diagnosis and downstream clinical analysis because lesion extent and boundary quality affect subsequent assessment, measurement, and intervention support.
Recent segmentation models such as PraNet and SegFormer have improved overall mask quality substantially, yet their predictions often remain locally imperfect around ambiguous boundaries, thin structures, and difficult coverage regions \cite{fan2020pranet,xie2021segformer,yue2024boundary}.
In colonoscopy images, these residual errors often appear near blurred lesion contours, low-contrast mucosal transitions, specular highlights, and small protruding structures that are visually subtle but still clinically relevant.
They are usually not distributed uniformly across the whole mask, but instead cluster in limited local regions that are harder to segment reliably.
This gap between globally reasonable masks and locally unreliable contours makes polyp segmentation a natural setting for lightweight post-refinement.

\begin{figure}[t]
\centering
\includegraphics[width=\linewidth]{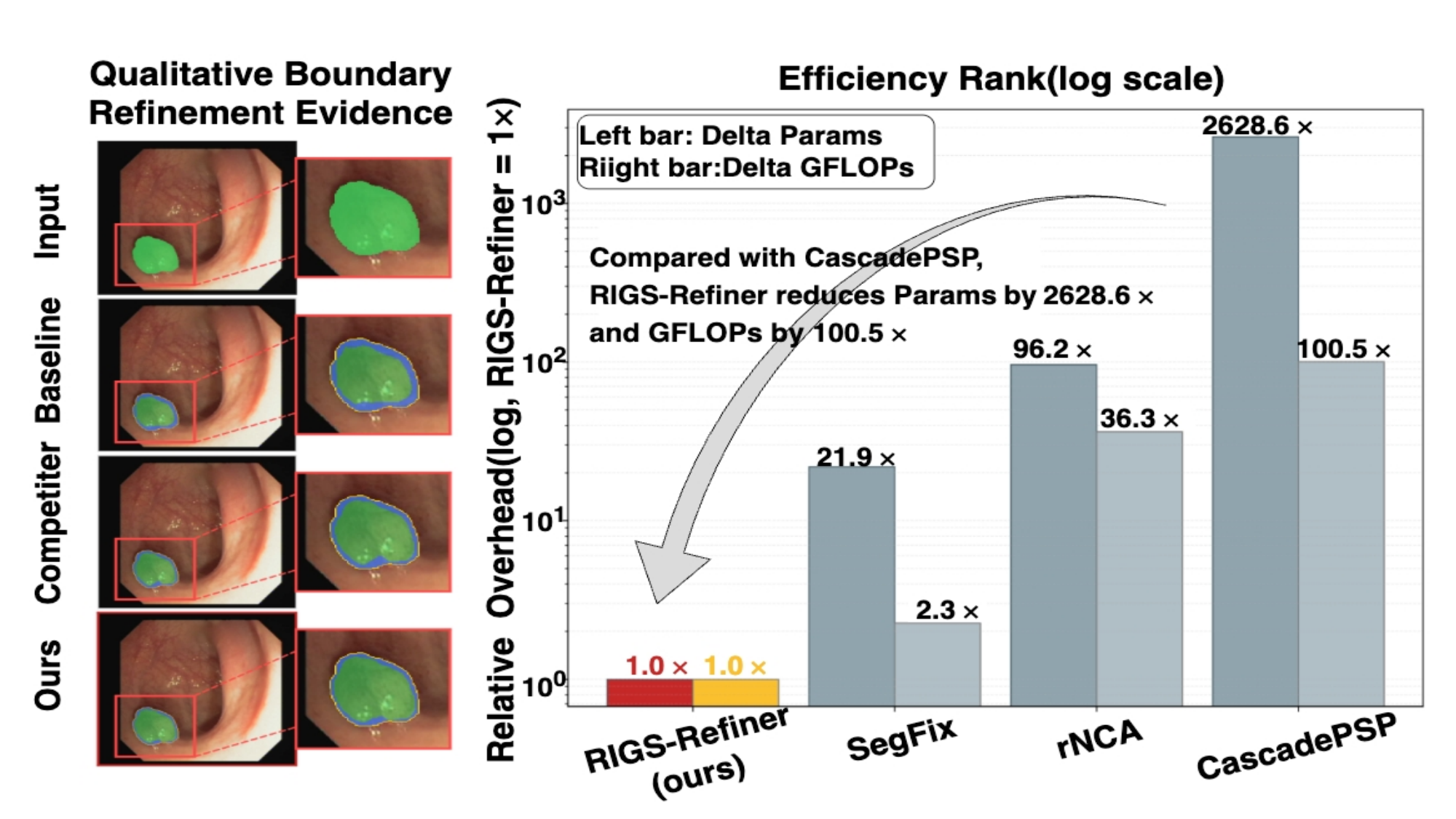}
\caption{Qualitative Local Refinement and Efficiency Comparison of RIGS-Refiner. Left: on a representative ClinicDB \cite{bernal2015wm} example, RIGS-Refiner produces more accurate local segmentation than the frozen host baseline and a representative post-refinement competitor. Right: relative refinement overhead on a log scale, where the left bar denotes added parameters and the right bar denotes added GFLOPs, both normalized by RIGS-Refiner ($1\times$). Compared with representative post-refinement alternatives, RIGS-Refiner keeps the refinement overhead much lower while remaining effective in local correction.}
\label{fig:intro_qual_cost}
\end{figure}

This makes post-refinement a practical direction, but not every refinement strategy is equally well matched to the problem considered here.
Existing methods explore several routes, including cascaded global-to-local correction \cite{cheng2020cascadepsp}, boundary-oriented refinement \cite{yuan2020segfix,borse2021inverseform}, point-wise high-resolution correction \cite{kirillov2020pointrend}, and more recent medical-image refinement designs \cite{zhang2024testfit,ke2025core,wu2024harmonizing,silbernagel2025rnca}.
While effective in many settings, these approaches often rely on additional refinement stages, extra correction heads, or more involved refinement pipelines.
For lightweight polyp segmentation, however, many difficult cases do not require the whole image to be decoded again, but rather a cleaner correction around a limited set of uncertain local regions.
Once the host prediction is already usable at the global level, reopening a heavy feature pathway can become inefficient relative to the scale of the remaining error.
The resulting question is therefore simple and task-specific: can local prediction errors in polyp segmentation be corrected effectively with a much smaller refinement mechanism that focuses on unreliable regions while leaving already reliable regions largely undisturbed?

In this paper, we propose \emph{RIGS-Refiner}, a lightweight post-refinement plugin for risk-guided recursive refinement in prediction space.
Starting from a frozen host anchor prediction, RIGS-Refiner extracts lightweight image priors and prediction cues, performs \emph{Risk-Guided Update}, and generates \emph{Residual Correction} within a shared recursive loop.
Rather than decoding a new segmentation mask from scratch, the design keeps the host prediction as an anchor and concentrates model capacity on selective local repair.
This formulation is well suited to the targeted deployment setting because it separates the effect of the refinement module from that of backbone redesign and keeps the added computation explicit.
As shown in Fig.~\ref{fig:intro_qual_cost}, the resulting design improves representative local segmentation results while adding only a very small refinement overhead.

The proposed design is lightweight, introducing only $+519$ parameters and $+0.631$ GFLOPs, which makes it well suited to deployment-oriented colonoscopy segmentation settings with varying image quality and acquisition conditions.
Experiments on Kvasir-SEG \cite{jha2019kvasir} and three out-of-distribution benchmarks, namely ClinicDB \cite{bernal2015wm}, ColonDB \cite{tajbakhsh2015automated}, and ETIS \cite{silva2014toward}, show that RIGS-Refiner improves the tested frozen host predictions under both PraNet and SegFormer-B0 backbones.
Compared with representative post-refinement methods such as SegFix, rNCA, and CascadePSP \cite{yuan2020segfix,silbernagel2025rnca,cheng2020cascadepsp}, it offers a favorable efficiency--accuracy trade-off with minimal added complexity.

In summary, this paper makes three contributions that address this task-specific question.
First, we formulate lightweight post-refinement for polyp segmentation as risk-guided recursive local repair in prediction space.
Second, we propose a compact refinement plugin that implements this formulation through image priors, prediction cues, risk-guided update, and residual correction within a shared recursive loop.
Third, we show that this design improves frozen host predictions across different hosts and datasets with very small parameter and computational overhead under a clear deployment budget.

\section{Related Work}
Recent progress in lightweight medical image segmentation has been driven largely by stronger host architectures, from classical encoder--decoder networks such as U-Net \cite{ronneberger2015u} to stronger hybrid variants such as TransUNet \cite{chen2021transunet}, PraNet \cite{fan2020pranet}, and SegFormer \cite{xie2021segformer}.
Within polyp segmentation in particular, many effective gains come from stronger encoders, richer decoder interactions, or host-specific architectural tailoring.
These contributions are important because they improve the quality of the host predictor itself.
The question studied here is different: rather than redesigning the predictor, we ask whether a lightweight correction module can be attached to a frozen host and still improve the final mask under a clear deployment budget.

Post-refinement has also been explored through several technical routes.
Classical post-processing methods such as DenseCRF \cite{krahenbuhl2011efficient} improve mask consistency through structured smoothing, while more recent approaches use cascaded refinement \cite{cheng2020cascadepsp}, point-based correction \cite{kirillov2020pointrend}, boundary-aware repair \cite{yuan2020segfix,borse2021inverseform}, and self-repairing iteration \cite{silbernagel2025rnca}.
Related medical-image designs further incorporate uncertainty cues, collaborative refinement, or boundary-oriented correction \cite{wang2020uncertainty,zhang2024testfit,ke2025core,wang2025dpgnet,gao2026mfbru}.
These directions are relevant to polyp segmentation because contour fidelity matters directly, but they still do not directly answer our target question.
Most are designed to increase correction capacity rather than to test how far a tiny external plugin can go under a strict lightweight budget.
What remains open is whether prediction-space refinement can stay effective when the design is constrained to be small, recursive, and explicitly attached to a frozen host.

\section{Method}
\subsection{Overview}
\begin{figure*}[t]
\centering
\includegraphics[width=\textwidth]{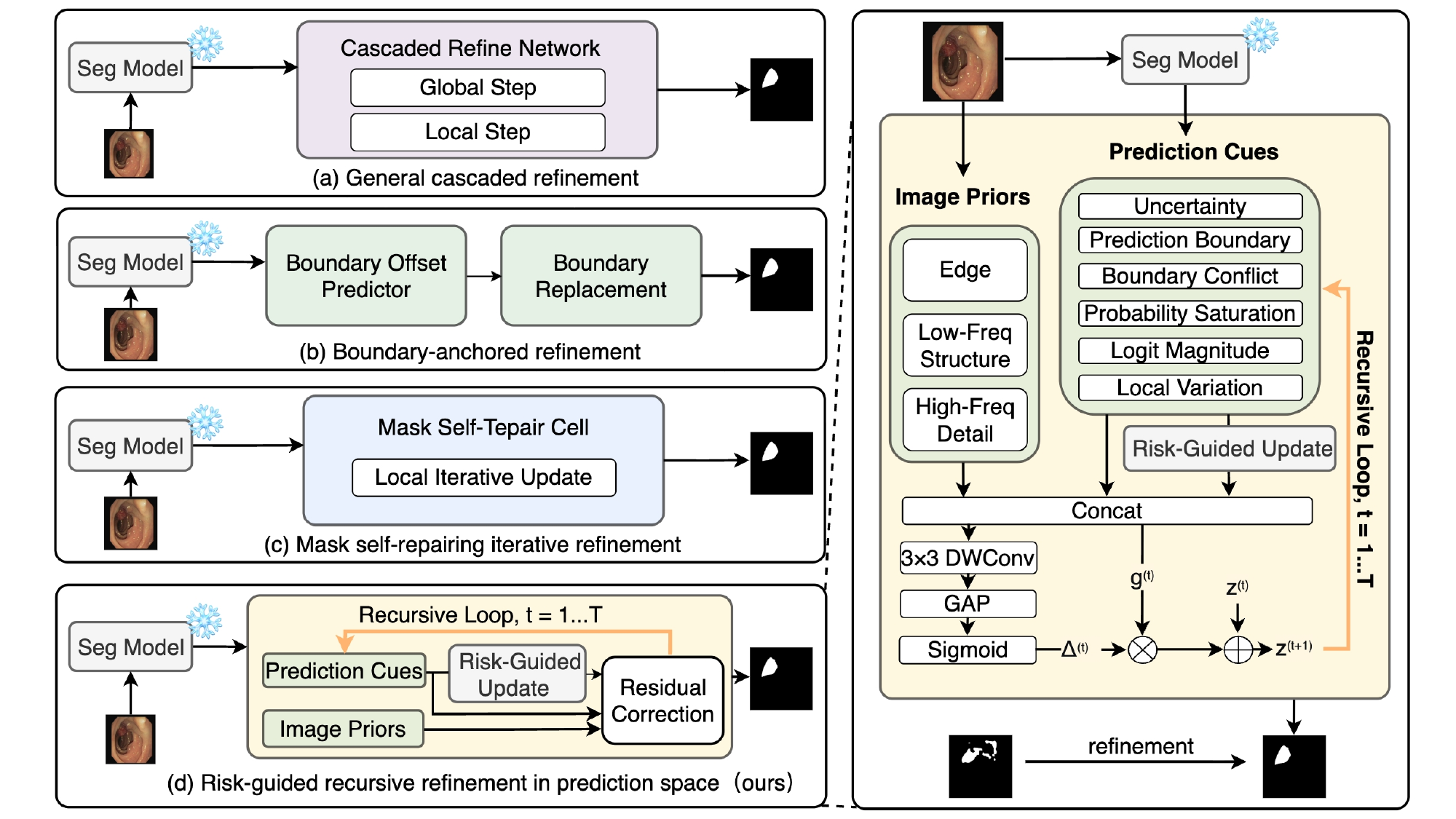}
\caption{Overview of Refinement Paradigms and RIGS-Refiner. Left: four representative refinement formulations are shown for comparison, including general cascaded refinement, boundary-anchored refinement, mask self-repairing iterative refinement, and our risk-guided recursive refinement in prediction space. Right: given a frozen host prediction anchor, RIGS-Refiner extracts image priors and prediction cues, applies risk-guided update, and generates residual corrections within a recursive loop over $t=1,\ldots,T$ to obtain the final refined mask. The snowflake symbol indicates the unified frozen-host training protocol shared across the compared refinement settings.}
\label{fig:method_overview}
\end{figure*}

RIGS-Refiner is a lightweight prediction-space post-refinement module for polyp segmentation, built on top of a frozen host segmenter.
Fig.~\ref{fig:method_overview} summarizes both representative refinement paradigms and the overall recursive pipeline of RIGS-Refiner.
Given an input image $x$, the host produces an anchor foreground logit map $z^{0}$ and its probability map
\begin{equation}
    p^{0} = \sigma(z^{0}),
\end{equation}
where $\sigma(\cdot)$ is the sigmoid function.
Instead of re-segmenting the whole image, RIGS-Refiner repeatedly applies a shared lightweight refinement cell to correct the anchor prediction only where it appears unreliable.
This design keeps the host output as a stable reference throughout the recursion, so the refinement path can focus on local error correction rather than global mask regeneration.

\subsection{Risk-Guided Recursive Refinement}
At recursive step $t$, the state is represented by foreground logit $z^{t}$ and probability $p^{t}=\sigma(z^{t})$.
The refinement cell combines image-side and prediction-side evidence to estimate where correction is needed and what should be corrected.
The right panel of Fig.~\ref{fig:method_overview} illustrates how image priors, grouped prediction cues, risk-guided update, and residual correction interact within this recursive process.
The cue boxes in the figure are schematic groups rather than a one-box-per-variable listing of all scalar elements in $q^{t}$.
At each step, the cell addresses two coupled questions: where the current mask remains uncertain, and what residual update should be written back there.

The image-side branch extracts three lightweight priors,
\begin{equation}
    e = \mathcal{N}(\mathcal{E}(x)), \qquad
    l = \mathcal{A}_{7}(x), \qquad
    h = x - l,
\end{equation}
where $\mathcal{E}(\cdot)$ denotes the Sobel edge operator, $\mathcal{A}_{7}(\cdot)$ denotes $7\times7$ average pooling, and $\mathcal{N}(\cdot)$ denotes per-image min-max normalization.
These priors are concatenated and projected into a compact image feature
\begin{equation}
    f^{\mathrm{img}} = \operatorname{Conv}([e,l,h]).
\end{equation}
This image-side feature is computed once and shared across recursive refinement steps.

The prediction-side branch builds a lightweight cue stack from the current prediction state.
These cues capture foreground probability, uncertainty, boundary response, image--prediction boundary discrepancy, confidence margin, residual magnitude, and local variation:
\begin{equation}
    u^{t} = - p^{t}\log p^{t} - (1-p^{t})\log(1-p^{t}),
\end{equation}
\begin{equation}
    b_{p}^{t} = \mathcal{N}(\mathcal{E}(p^{t})), \qquad
    c^{t} = \left| \mathcal{N}(e) - b_{p}^{t} \right|,
\end{equation}
\begin{equation}
    s^{t} = 2\left| p^{t} - 0.5 \right|, \quad
    a^{t} = \mathcal{N}(|z^{t}|), \quad
    v^{t} = \mathcal{N}\!\left( \left| z^{t} - \mathcal{A}_{7}(z^{t}) \right| \right).
\end{equation}
Together, they provide a compact view of the current prediction, its uncertainty and boundary structure, its agreement with the image-side boundary prior, and the strength and local variation of the current logit state.
They are concatenated into
\begin{equation}
    q^{t} = [p^{t}, u^{t}, b_{p}^{t}, c^{t}, s^{t}, a^{t}, v^{t}],
\end{equation}
and compressed into a prediction feature
\begin{equation}
    f^{\mathrm{pred}}_{t} = \operatorname{Conv}(q^{t}),
\end{equation}
followed by a risk score
\begin{equation}
    r^{t} = \operatorname{Conv}(f^{\mathrm{pred}}_{t}), \qquad
    g^{t} = \sigma(r^{t}),
\end{equation}
where $g^{t}$ serves as a soft update mask.
In this paper, Risk-Guided Update is therefore interpreted as a learned selective update control that modulates where and how strongly residual corrections are written back.

The correction core predicts a residual logit update from the image feature, prediction feature, and gate:
\begin{equation}
    h^{t} = \operatorname{Conv}([f^{\mathrm{img}}, f^{\mathrm{pred}}_{t}, g^{t}]),
\end{equation}
\begin{equation}
    \hat{h}^{t} = h^{t} + \mathrm{DWConv}_{3\times3}(h^{t}),
\end{equation}
\begin{equation}
    \Delta^{t} = \operatorname{Conv}\!\left(
    \operatorname{ReLU}\!\left(
    \hat{h}^{t} \odot \sigma(\mathrm{GAP}(\hat{h}^{t}))
    \right)\right).
\end{equation}
The final update is then written back as
\begin{equation}
    z^{t+1} = z^{t} + g^{t} \odot \Delta^{t}, \qquad
    p^{t+1} = \sigma(z^{t+1}).
\end{equation}
The recursion stops at $t=T$, yielding the final refined prediction.
Because the same cell is reused across steps, recursive depth increases correction capacity without introducing separate stage-specific refinement heads.

\subsection{Training Objective}
Training uses the segmentation loss on the final refined prediction:
\begin{equation}
    \mathcal{L} = \mathcal{L}_{\mathrm{Dice}}(p^{T}, y),
\end{equation}
where $y$ is the ground-truth mask.
This objective keeps optimization simple and aligned with the module role, refining the final prediction without an additional supervision-heavy correction branch or auxiliary step-specific losses.

\section{Experiments}
\subsection{Experimental Setup}
All datasets used in this work are polyp segmentation benchmarks.
We use Kvasir-SEG \cite{jha2019kvasir} as the training dataset and evaluate generalization on both the in-domain Kvasir split \cite{jha2019kvasir} and three out-of-distribution polyp datasets, namely ClinicDB \cite{bernal2015wm}, ColonDB \cite{tajbakhsh2015automated}, and ETIS \cite{silva2014toward}.
OOD evaluation is conducted by directly applying the best Kvasir-trained checkpoint to the full target dataset without any target-domain finetuning.
This evaluation is important for polyp segmentation because lesion appearance, image quality, and acquisition conditions vary substantially across datasets and clinical centers, which makes robustness to medical image signal variation particularly relevant.

We consider two host backbones, namely PraNet and SegFormer-B0 \cite{fan2020pranet,xie2021segformer}.
These hosts provide complementary CNN-based and Transformer-based predictors for testing whether the same refinement plugin remains useful across different frozen prediction styles.
Each host is first trained on Kvasir-SEG for 200 epochs using AdamW with an initial learning rate of $1\times10^{-4}$ and cosine learning-rate decay.
Both hosts are trained with Dice-based objectives.
Input images are resized to $352\times352$, and both training and validation batch sizes are set to 16.
After host training, the best host checkpoint is frozen and used as the initialization anchor for post-refinement training.
RIGS-Refiner is then trained for 50 epochs on Kvasir-SEG with AdamW, an initial learning rate of $1\times10^{-4}$, cosine learning-rate decay, the same input resolution, and the same batch size.
For competing post-refinement methods, we follow their original training objectives and faithful implementation protocols rather than forcing a unified optimization setting across methods.
All training is implemented in PyTorch and conducted on a single NVIDIA A100 80G GPU.
This protocol keeps the comparison centered on post-refinement behavior rather than on differences caused by host retraining or target-domain adaptation.
Model complexity is profiled with THOP, and the reported values follow the adopted THOP convention.

Following recent post-refinement evaluation practice \cite{silbernagel2025rnca}, we report IoU, HD95, and clDice in the main table.
IoU measures region overlap, HD95 reflects boundary robustness under the 95th-percentile Hausdorff distance, and clDice evaluates structural connectivity.

\subsection{Comparison with Existing Methods}
\paragraph{Broad Gains with Minimal Overhead.} Table~\ref{tab:merged_main} compares RIGS-Refiner with the frozen host baseline and three representative post-refinement methods, namely SegFix, rNCA, and CascadePSP. Under PraNet, RIGS-Refiner improves over the frozen baseline on all four datasets, with IoU gains of $+0.68$ on Kvasir, $+0.57$ on K$\rightarrow$Clinic, $+0.85$ on K$\rightarrow$Colon, and $+0.67$ on K$\rightarrow$ETIS, while also reducing HD95 by $0.77$, $4.06$, $3.78$, and $5.56$, respectively. Under SegFormer-B0, the gains are smaller but still consistent, with IoU improvements of $+0.27$, $+0.22$, $+0.15$, and $+0.55$ across the same four datasets and corresponding HD95 reductions of $0.62$, $0.10$, $0.18$, and $5.08$. Against competing post-refinement methods, RIGS-Refiner does not dominate every single metric, but it remains competitive across the full comparison, matches or exceeds the compared methods on several dataset--metric pairs, and does so while adding only $+519$ parameters and $+0.631$ GFLOPs, compared with $+11{,}369$ and $+1.42$ for SegFix, $+13{,}649$ and $+3.78$ for rNCA, and $+1{,}364{,}265$ and $+63.46$ for CascadePSP. The main result is therefore not a uniform win on all entries, but a notably strong efficiency--accuracy trade-off under a very small refinement budget.

\begin{table*}[t]
\centering
\caption{Quantitative Comparison on Four Datasets Under Two Backbones. Bold indicates the best result, and underline indicates the second best. $\uparrow$: higher is better; $\downarrow$: lower is better.}
\label{tab:merged_main}
\resizebox{\textwidth}{!}{
\begin{tabular}{ll|ccc|ccc|ccc|ccc|rr}
\toprule
\multirow{2}{*}{Backbone} & \multirow{2}{*}{Method}
 & \multicolumn{3}{c|}{Kvasir}
 & \multicolumn{3}{c|}{K$\to$Clinic}
 & \multicolumn{3}{c|}{K$\to$Colon}
 & \multicolumn{3}{c|}{K$\to$ETIS}
 & \multirow{2}{*}{$\Delta$Params} & \multirow{2}{*}{$\Delta$GFLOPs} \\
\cmidrule{3-14}
 &  & IoU$\uparrow$ & HD95$\downarrow$ & clDice$\uparrow$
    & IoU$\uparrow$ & HD95$\downarrow$ & clDice$\uparrow$
    & IoU$\uparrow$ & HD95$\downarrow$ & clDice$\uparrow$
    & IoU$\uparrow$ & HD95$\downarrow$ & clDice$\uparrow$ & & \\
\midrule
\multirow{5}{*}{PraNet}
 & Baseline             & 85.66 & 20.21 & 92.33 & 71.75 & 67.93 & 78.86 & 41.91 & 156.58 & 60.12 & 65.57 & 234.24 & 51.00 & 0 & 0.000 \\
 & SegFix               & 86.32 & 19.45 & 93.20 & 72.23 & 64.21 & 79.30 & 42.39 & 153.03 & 60.94 & 66.16 & 230.42 & 52.37 & +11\,369 & +1.42 \\
 & rNCA                 & \second{86.33} & \best{19.39} & 93.25 & 72.27 & \second{64.10} & 79.51 & \second{42.58} & \second{152.95} & \second{61.12} & \best{66.31} & \second{230.35} & \second{52.40} & +13\,649 & +3.78 \\
 & CascadePSP           & 86.32 & 19.62 & \best{93.46} & \best{72.33} & 64.93 & \second{79.53} & 42.45 & 154.21 & 61.11 & 66.23 & 232.37 & \best{52.52} & +1\,364\,265 & +63.46 \\
 & RIGS-Refiner (Ours)  & \best{86.34} & \second{19.44} & \second{93.32} & \second{72.32} & \best{63.87} & \best{79.64} & \best{42.76} & \best{152.80} & \best{61.27} & \second{66.24} & \best{228.68} & 52.39 & \best{+519} & \best{+0.631} \\
\midrule
\multirow{5}{*}{SegFormer-B0}
 & Baseline             & 84.90 & 21.37 & 90.79 & 75.07 & 31.46 & 84.39 & 57.68 & 58.93 & 74.36 & 67.52 & 67.17 & 71.80 & 0 & 0.000 \\
 & SegFix               & 85.12 & 20.76 & 91.16 & 75.28 & \second{31.38} & 84.80 & 57.71 & 58.95 & \second{74.50} & \second{68.04} & \second{64.45} & \second{72.08} & +11\,369 & +1.42 \\
 & rNCA                 & 85.15 & \best{20.66} & 91.01 & \best{75.31} & 31.39 & 84.76 & 57.70 & \second{58.83} & 74.29 & 68.01 & 64.54 & 71.96 & +13\,649 & +3.78 \\
 & CascadePSP           & \best{85.18} & 20.74 & \second{91.47} & 75.25 & 64.93 & \best{85.14} & \best{57.92} & \best{58.57} & \best{75.49} & 67.89 & 67.07 & \best{72.85} & +1\,364\,265 & +63.46 \\
 & RIGS-Refiner (Ours)  & \second{85.17} & \second{20.75} & \best{91.59} & \second{75.29} & \best{31.36} & \second{84.91} & \second{57.83} & 58.75 & 74.75 & \best{68.07} & \best{62.09} & 71.97 & \best{+519} & \best{+0.631} \\
\bottomrule
\end{tabular}}
\end{table*}

\subsection{Ablation Studies}
We conduct two focused ablation studies under the PraNet backbone.
PraNet is used here as a representative host for mechanism analysis, while the main comparison across PraNet and SegFormer-B0 already establishes the plugin-level effectiveness under two different frozen prediction styles.
The first study examines the loop budget $T$ to test whether recursive refinement is genuinely more effective than a single correction pass.
The second study examines the contribution of the two core mechanisms in the full model, namely deep recurrence and risk-guided update.

\begin{table}[t]
\centering
\caption{Ablation on Loop Budget $T$ Under PraNet}
\label{tab:ablation_T}
\begin{tabular}{lrrr}
\toprule
Loop Budget $T$ & IoU(\%)$\uparrow$ & HD95$\downarrow$ & clDice$\uparrow$ \\
\midrule
PraNet Baseline ($T=0$) & 85.658 & 20.213 & 92.330 \\
$T=1$ (Ours)            & 86.222 & 19.473 & 93.263 \\
$T=2$ (Ours)            & \second{86.328} & \best{19.435} & \second{93.279} \\
$T=3$ (Ours)            & \best{86.344} & \second{19.437} & \best{93.322} \\
\bottomrule
\end{tabular}
\end{table}

\paragraph{Recursive Depth Brings Incremental Benefit.} The loop-budget results show that the benefit of refinement is not exhausted after a single update. Relative to the frozen PraNet baseline, all tested loop budgets improve IoU, HD95, and clDice. Moving from $T=1$ to $T=3$ further raises IoU from $86.222$ to $86.344$ and clDice from $93.263$ to $93.322$, while HD95 remains at a similarly improved level, changing from $19.473$ to $19.437$. These gains are modest, but they are consistent with the intended role of recursion here: when the host prediction is already reasonable at the global level, several lightweight correction steps can gradually repair residual local errors that are difficult to remove in a single pass.

\begin{table}[t]
\centering
\caption{Component-Wise Ablation on PraNet}
\label{tab:ablation_component}
\begin{tabular}{lrrr}
\toprule
Setting & IoU(\%)$\uparrow$ & HD95$\downarrow$ & clDice$\uparrow$ \\
\midrule
Backbone only               & 85.658 & 20.213 & 92.330 \\
w/o Risk-Guided Update      & 86.252 & 19.610 & 92.972 \\
w/o Deep Recurrence ($T=1$) & 86.222 & 19.473 & 93.263 \\
Full (Ours)                 & \best{86.344} & \best{19.437} & \best{93.322} \\
\bottomrule
\end{tabular}
\end{table}

\paragraph{Both Core Mechanisms Contribute.} The component ablation shows that each part of the full design adds measurable value. Removing Risk-Guided Update still improves over the backbone-only baseline, which indicates that image priors and residual correction already provide useful local repair information. However, the full model remains better on all three metrics, improving over the ungated variant by $+0.092$ IoU, reducing HD95 by $0.173$, and increasing clDice by $0.350$. Within this evidence, Risk-Guided Update should be read as a selective update control that contributes additional value beyond an ungated correction path, rather than as a fully isolated proof of spatial risk estimation. A similar pattern appears when the full model is compared with the $T=1$ variant: the gains of $+0.122$ IoU, $0.036$ lower HD95, and $+0.059$ clDice are small but consistent, which suggests that deep recurrence is not just a repeated application of the same correction step, but a mechanism that allows local updates to accumulate more effectively.

\section{Discussion}
Quantitative results show that RIGS-Refiner improves the tested frozen host predictions across multiple datasets and backbones while adding only $+519$ parameters and $+0.631$ GFLOPs.
Once the frozen host already provides a reasonable global prediction anchor, the remaining problem is much narrower than full segmentation: the main challenge is to identify unreliable local regions and correct them without disturbing areas that are already correct.
In polyp segmentation, this setting is particularly common because difficult cases are often defined less by complete target omission than by incomplete lesion coverage, contour leakage, or missed thin boundary fragments.

Rather than learning another dense prediction head, RIGS-Refiner divides the refinement problem into complementary roles: lightweight image priors provide structural evidence from the input image, prediction cues describe the reliability of the current prediction, and risk-guided update controls where corrective residuals are written back.
Model capacity is thus spent on error localization and selective correction instead of redundant full-mask reconstruction.
Taken together, the current evidence supports a narrow claim: the method can improve frozen host predictions while keeping the refinement budget extremely small.
It is therefore best positioned as a complementary refinement layer atop existing host architectures, rather than as a replacement for stronger host redesign or heavier multi-stage refinement.
This also fits practical colonoscopy deployment, where lightweight adaptation of an existing host can be more realistic than retraining or replacing the full system.

Another pattern is that the gains appear in both overlap-oriented and boundary-oriented metrics rather than in only one view of quality.
This matters for polyp segmentation because practical mask quality depends not only on coarse lesion coverage, but also on how well the predicted contour follows the true lesion extent.
The improvements in IoU, HD95, and clDice therefore suggest that the proposed plugin is not merely smoothing predictions, but improving mask quality from multiple structural perspectives.
Just as importantly, these gains are obtained without relying on a large refinement branch, so the method remains attractive even in cases where some competing methods can still be stronger on individual entries.

The contribution of the present study should be understood within lightweight post-refinement on top of frozen hosts.
The method is not intended to replace stronger host redesign or outperform every heavier refinement method in absolute performance.
Instead, a very small prediction-space correction module remains competitive while keeping the added cost explicit and tightly bounded.

This makes the method especially relevant when a reliable host is already available and full pipeline redesign is unnecessary or too costly.
In practical colonoscopy systems, refinement often needs to be added to an established pipeline without materially changing its parameter budget, training protocol, or inference flow, so the method is best regarded as a lightweight correction path attached to an existing predictor rather than as a universal substitute for architectural improvement.
Future work can examine how the same prediction-space refinement principle transfers to broader host families, more diverse acquisition conditions, and refinement settings with richer image-side evidence.

\section{Conclusion}
We presented RIGS-Refiner, a lightweight post-refinement plugin for colonoscopy polyp segmentation that performs risk-guided recursive refinement directly in prediction space.
Built on a frozen host prediction anchor, the method combines image priors, prediction cues, risk-guided update, and residual correction within a shared recursive cell.
Experiments on PraNet and SegFormer-B0 across Kvasir, ClinicDB, ColonDB, and ETIS show improved frozen host predictions with only $+519$ parameters and $+0.631$ GFLOPs, keeping the refinement path compact and deployment-friendly.
These results support the practical point that, for frozen host predictions in polyp segmentation, useful refinement need not require another heavy segmentation branch.
Under this setting, prediction-space recursion with selective write-back efficiently corrects local mask errors while keeping the added cost small.

\bibliographystyle{unsrtnat}
\bibliography{ref}

\end{document}